\documentclass[pdflatex,sn-mathphys-num]{sn-jnl}

\usepackage{graphicx}%
\usepackage{multirow}%
\usepackage{amsmath,amssymb,amsfonts}%
\usepackage{amsthm}%
\usepackage{mathrsfs}%
\usepackage[title]{appendix}%
\usepackage{xcolor}%
\usepackage{textcomp}%
\usepackage{manyfoot}%
\usepackage{booktabs}%
\usepackage{algorithm}%
\usepackage{algorithmicx}%
\usepackage{algpseudocode}%
\usepackage{listings}%
\usepackage{booktabs}
\usepackage{siunitx}
\usepackage{cleveref}

\theoremstyle{thmstyleone}%
\theoremstyle{thmstyletwo}%

\theoremstyle{thmstylethree}%

\begin{document}

\title[Article Title]{Beyond Thermal Imaging: Inferring Thermophysical Properties from Time-Resolved Thermal Observations}


\author[1]{\fnm{Chenghao} \sur{Xu}}\email{chenghao.xu@epfl.ch}

\author[2]{\fnm{Malcolm} \sur{Mielle}}\email{malcolm.mielle@ik.me}

\author*[1]{\fnm{Olga} \sur{Fink}}\email{olga.fink@epfl.ch}
\affil[1]{\orgdiv{Intelligent Maintenance and Operations Systems}, \orgname{École polytechnique fédérale de Lausanne}, \orgaddress{\street{Station 18}, \city{Lausanne}, \postcode{1015}, \state{Vaud}, \country{Switzerland}}}

\affil[2]{\orgdiv{Schindler Lab EPFL}, \orgname{Schindler Holding AG}, \orgaddress{\street{Station 18}, \city{Lausanne}, \postcode{1015}, \state{Vaud}, \country{Switzerland}}}




\abstract{Inferring latent physical properties from sensory observations is a fundamental challenge in machine perception. Among available sensing modalities, thermal imaging is particularly promising because temperature evolution is directly governed by heat-transfer physics and therefore encodes information about the underlying thermophysical properties of a scene. Recovering spatially resolved thermophysical properties from thermal observations could transform applications ranging from digital twins and infrastructure monitoring to robotics and scientific imaging. However, existing thermal scene reconstruction methods can recover temperature fields in complex three-dimensional environments without identifying the underlying thermophysical properties that govern thermal evolution, whereas inverse heat-transfer methods provide physically interpretable parameter estimation but typically rely on simplified geometries and controlled experimental conditions. Consequently, recovering spatially varying thermophysical properties in realistic scenes remains an open challenge.

Here we introduce \textbf{ThermoField}, a framework that unifies thermal scene reconstruction and thermophysical parameter estimation through differentiable heat-transfer simulation. Rather than treating temperature as a visual attribute to be reconstructed, we interpret thermal observations as measurements of an underlying heat-transfer process. The proposed framework represents thermophysical quantities as spatially varying neural fields and constrains them through scene geometry, governing heat-transfer physics, and temporal thermal observations. This formulation makes thermophysical parameters identifiable from their influence on observed thermal dynamics while remaining consistent with both measurements and physical laws.

We demonstrate that \textbf{ThermoField} jointly reconstructs geometry, estimates spatially varying thermal diffusivity and boundary heat-transfer coefficients, and predicts thermal evolution under previously unseen environmental conditions. By integrating neural scene representations with differentiable heat-transfer physics, the framework enables physically interpretable thermophysical inference in complex three-dimensional scenes. Our results establish a bridge between thermal scene reconstruction and inverse heat-transfer analysis, providing a unified approach for geometry reconstruction, thermophysical property estimation, and predictive thermal simulation from thermal observations.

}




\maketitle

\section{Introduction}
\label{sec:intro}
A longstanding goal of artificial intelligence is to recover the latent physical properties that govern how the world behaves, rather than merely describing how it appears. Humans routinely infer material composition, thermal behavior, and physical dynamics from observation, enabling prediction, intervention, and planning. Equipping machine perception with similar capabilities would establish a foundation for physically grounded AI systems that can reason about, simulate, and interact with the physical world. Such physically interpretable representations would support predictive simulation~\cite{PhysAavatar24, Li_2025_CVPR}, counterfactual reasoning~\cite{10.5555/3666122.3669051, chow2025physbench}, digital twins~\cite{zhou2026digitaltwinaiopportunities}, and autonomous interaction with complex environments~\cite{10.1007/978-3-031-99565-1_8, Chen24physical}. Together, these capabilities make the recovery of physical properties from visual observations a central challenge at the intersection of computer vision, scientific machine learning, and physical AI.

Despite remarkable advances in visual perception, existing vision systems remain largely appearance-driven. Although highly successful at recognition and geometric reconstruction, they typically estimate approximations of physical properties only indirectly, by learning statistical correlations between visual appearance and  material or physical attributes~\cite{pgvlm2024, zhao2024efficient}, rather than explicitly reasoning about the underlying physical processes.
Such associations are effective for semantic perception, but provide only weak evidence of the underlying physical process because visual appearance is shaped by multiple interacting factors, including material composition, geometry, illumination, and sensor characteristics.
Consequently, recovering physical properties from visual observations is intrinsically ill-posed. Visually similar objects may exhibit fundamentally different physical behavior, whereas physically similar systems may appear visually distinct. This disconnect limits the ability of current AI systems to support physically grounded prediction, simulation, and scientific inference, motivating representations that recover the latent physical processes governing the observed scene.

In conventional RGB imagery, this challenge is particularly severe. Cues related to material composition, heat transfer, and scene dynamics are deeply entangled with geometry, illumination, viewpoint, and sensor effects, making the inverse problem severely underconstrained~\cite{bryan1996inverse, dashpute2023thermal, https://doi.org/10.1111/j.1467-8659.2003.00716.x}.
Additional observations, including multi-view images~\cite{Li_NeISF_CVPR2024, wu2025pbrnerf}, temporal video~\cite{humanMotionKZFM19, chen2025vid2sim}, or geometric information such as depth and surface normals~\cite{10.1007/978-3-030-58583-9_1}, can partially alleviate these ambiguities.
However, they primarily improve geometric observability, rather than directly revealing the latent physical parameters governing material behavior. Reliable estimation of physically interpretable properties therefore remains an open challenge.
This motivates the exploration of sensing modalities whose temporal evolution more directly encodes the physical processes governing the scene.

Thermal imaging provides a unique opportunity to bridge perception and physics. Unlike RGB cameras, thermal cameras measure radiometric emission in the long-wave infrared spectrum, providing observations that are closely related to surface temperature~\cite{physics_of_thermal}.
Crucially, temperature evolution is governed by the laws of heat transfer. Time-resolved thermal measurements therefore encode the spatiotemporal evolution of heat as it diffuses through materials and across interfaces, providing observations of an underlying physical state rather than purely visual appearance~\cite{ch4}. 
In principle, these dynamics contain information about thermophysical and interfacial quantities, such as thermal diffusivity, surface emissivity, and heat-transfer coefficients. Thermal imagery therefore represents a rare sensing modality whose measurements are intrinsically coupled to governing physical processes, offering stronger guidance on latent physical properties than conventional appearance-based observations.

However, more physically informative measurements alone do not resolve the inverse problem. The central challenge is physical identifiability: the observed temperature field is jointly shaped by scene structure, material properties, boundary conditions, and heat exchange with the environment. 
Through the heat equation, different combinations of these factors can generate similar surface temperature evolution, making the underlying parameters difficult to disentangle.
As a result, recovering geometry and thermophysical parameters from thermal observations remains highly ambiguous, especially under limited viewpoints, partial observations, or short temporal windows~\cite{10.1117/12.3065037, Narayanan_2026_CVPR}. 
Without additional physical constraints or prior knowledge, these parameters may remain non-identifiable from thermal observations alone.

Existing thermal vision methods have largely approached thermal imagery from a perception perspective, leaving this identifiability challenge unaddressed. Earlier work focused on cross-modal detection~\cite{WANG2023105640}, segmentation~\cite{9108585}, and anomaly localization~\cite{vollmer2024detecting}. In these settings, thermal data are used as a complementary modality to RGB, improving robustness under poor illumination or adverse environmental conditions.
Although effective for perception, these approaches primarily treat thermal imagery as an image-level cue for feature fusion and semantic prediction, rather than as measurements of a physical state.

More recent work has extended thermal vision to 3D scene reconstruction, using neural radiance fields and Gaussian splatting techniques to recover geometrically consistent thermal maps or implicit temperature fields from image sequences~\cite{hassan2025thermonerf,chen2024thermal3dgs}.
Because thermal imagery often provides weak geometric texture, these methods typically rely on RGB images or LiDAR measurements to estimate geometry and camera motion before fusing thermal observations into the reconstructed scene.
Despite these advances in geometric reconstruction, temperature is generally modeled as a static appearance attribute attached to geometry, rather than as a dynamic physical field governed by heat transfer. Consequently, existing methods can reconstruct thermally consistent scenes, but remain limited in inferring thermophysical properties or predicting temperature evolution under previously unseen boundary conditions.

Beyond reconstructing thermal appearance, a complementary direction for physical scene analysis is physics-guided parameter estimation, where governing models are embedded into the optimization process to recover latent system parameters from visual observations~\cite{degrave2019differentiable, waseem2025physicsinformedneuralnetworksthermophysical}.
Such formulations provide a useful foundation for physically constrained estimation when a differentiable simulator or physical model is available.
This strategy has been primarily explored in deformable object modeling and dynamical system identification~\cite{hu2019difftaichi,NIPS2015_d09bf415}, where material or mechanical parameters, such as elasticity, stiffness, or damping, are estimated by matching observed motion or shape changes to model-based predictions.
Rather than regressing parameters directly from visual appearance, these approaches constrain inference through explicit dynamics, producing estimates that are more interpretable and physically consistent.

However, 
the effectiveness of this paradigm depends on whether the parameters are identifiable from the available observations and physical constraints.
In underconstrained vision-based settings, multiple parameter configurations may produce similar motion or deformation patterns, allowing model predictions to align closely with observations without uniquely recovering the true physical quantities of the real system. 
Consequently, accurately matching observations does not necessarily imply physically meaningful or transferable parameter estimation.
This limitation is particularly pronounced in thermophysical inference, where the target parameters are reflected only indirectly through temperature evolution and are strongly coupled with geometry, environmental conditions, and boundary interactions, rather than being revealed through explicit deformations, trajectories, or contacts.

Closer to thermophysical inference, the most relevant prior work lies in two directions: dynamic thermal scene representations and inverse heat-transfer methods.
Dynamic thermal scene models extend neural fields with time-varying temperature fields and physically motivated variables, such as emissivity, heat capacity, and heat-transfer coefficients~\cite{Yang_2025_CVPR, wangetgs}.
These formulations improve thermal view synthesis and temporal reconstruction, but their primary objective remains the high-fidelity rendering of observed thermal sequences.
As a result, the recovered parameters often serve as latent variables that explain observed temperature histories, rather than as identifiable thermophysical quantities that generalize across environmental conditions or support predictive reasoning.

As a thermophysical instance of physics-guided parameter estimation, inverse heat-transfer methods embed heat-transfer models into the inference process to recover physical properties from temperature measurements.
By constraining measured thermal transients with the heat equation, existing approaches have shown that heating and cooling responses can reveal material-dependent quantities, such as thermal diffusivity or emissivity~\cite{dashpute2023thermal, 7299096}.
However, these quantities are typically identifiable only under controlled assumptions, including prescribed excitation, simplified geometries, and constrained boundary conditions. Many existing formulations further discretize heat diffusion on regular grids or voxelized domains, where the governing spatial operator is approximated by local finite-difference stencils.
While effective for canonical inverse problems, these assumptions limit scalability to complex 3D scenes with irregular geometry, heterogeneous materials, and realistic environmental interactions.

Thermal scene reconstruction and inverse heat-transfer methods therefore address complementary aspects of thermophysical inference.
The former is well suited to complex 3D scenes, but typically focuses on recovering temperature fields or thermal appearance, without establishing physically meaningful parameters.
The latter provides physically interpretable parameter estimation, but usually relies on simplified geometries, constrained boundary conditions, or grid-based  diffusion operators that limit its applicability to realistic scenes.
Neither direction currently combines the ability to model complex 3D scenes with the recovery of thermophysical properties that remain predictive under changing environmental conditions.
Addressing this problem requires a differentiable scene  representation that couples geometry, surface thermal observations, and  heat-transfer dynamics, allowing spatially varying thermophysical properties to be inferred through physical constraints rather than treated merely as latent variables for reproducing temperature histories.

\begin{figure}[h]
\centering
\includegraphics[width=1.0\textwidth]{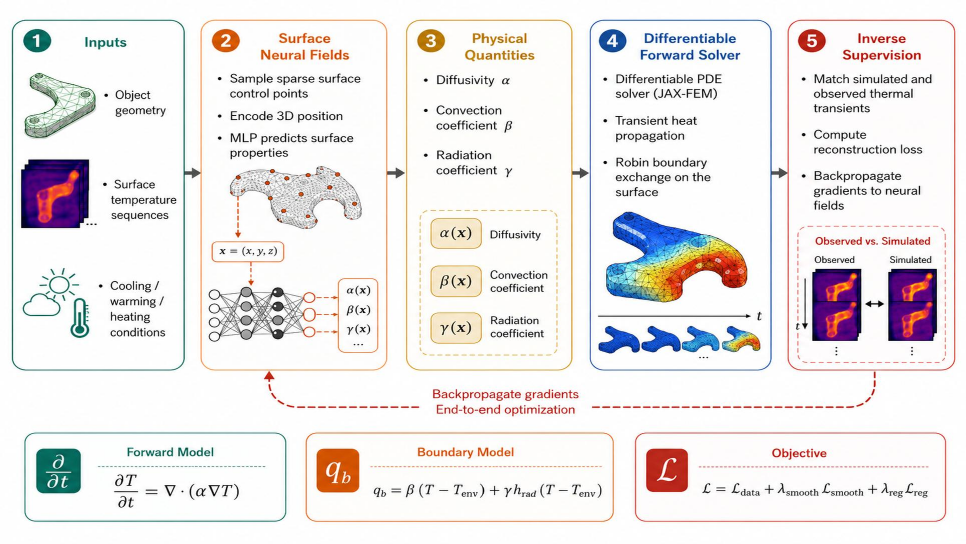}
\caption{ThermoField overview. ThermoField couples surface neural fields with differentiable transient heat-transfer simulation to infer thermophysical properties from surface thermal observations. The framework takes reconstructed object geometry, surface temperature sequences, and external heating or cooling conditions as input. Neural fields defined on sparse surface control points parameterize latent physical quantities, including thermal diffusivity, convective exchange, and radiative exchange coefficients. These quantities drive a differentiable forward solver that propagates heat diffusion and surface boundary exchange to produce simulated thermal transients. The simulated surface temperatures are matched to observed thermal sequences, and the resulting loss is backpropagated through the solver to update the neural physical fields. }
\label{fig:architecture}
\end{figure}

In this work, we introduce ThermoField, a physics-grounded framework for inferring spatially varying thermophysical fields on reconstructed 3D objects from time-resolved thermal observations, as shown in Fig.~\ref{fig:architecture}.
Unlike existing approaches that either reconstruct thermal appearance or solve inverse heat-transfer problems on simplified computational domains, ThermoField extends thermophysical inversion to geometrically complex reconstructed objects without simplifying the underlying object geometry.
The framework separates scene reconstruction from thermophysical inversion:
object geometry is first reconstructed from multi-view observations and used as a fixed computational domain for subsequent physical inference.
The observed thermal sequence is then registered onto the reconstructed surface, where spatially varying thermophysical properties are represented as neural fields defined on the object geometry.
These fields can parameterize heat-transfer-related quantities, including thermal diffusivity, convective exchange, and radiative exchange coefficients, which are coupled to a differentiable transient heat-transfer model for physics-constrained inference.

ThermoField treats thermal images not as appearance to be reproduced, but as surface measurements generated by an underlying heat-transfer process. This formulation exposes an intrinsically underconstrained inverse problem: thermal cameras observe only surface temperature, whereas internal temperature evolution and the physical parameters governing heat propagation remain hidden.
In the diffusivity-estimation setting considered in this work, the problem is further complicated by the small scale and broad range of thermal diffusivity, spanning approximately $10^{-7}$ to $10^{-4}$ $\mathrm{m^2\,s^{-1}}$.
Consequently, the recovered values can be highly sensitive to uncertainties  in geometry reconstruction, boundary modeling, temporal alignment, and temperature calibration.
Thermophysical fields are therefore inferred by requiring the forward heat-transfer model to match the observed surface temperature evolution while remaining consistent with the governing heat-transfer physics.

To make this inverse formulation operational, ThermoField represents heat-transfer-related quantities as explicit spatial fields on the reconstructed object. These fields can describe thermal diffusivity, convective-exchange coefficients, or radiative-exchange coefficients that enter the heat-transfer model.
In the diffusivity-estimation setting studied here, the convective and radiative terms are fixed, so that the inverse problem focuses on recovering the spatial distribution of thermal diffusivity. The diffusivity field is parameterized through sparse surface control points and a neural representation with positional encoding, and is extended into the volume when required by the forward simulation.
Rather than reducing the object to a single global material constant, this spatially resolved formulation can represent  local variations associated with  heterogeneous materials, surface treatments, geometric irregularities, or region-dependent thermal responses, while also enabling regional or global statistical analysis of the recovered thermophysical fields.

The key computational ingredient is that the forward heat-transfer process defines a differentiable map from physical fields to temperature evolution. After discretization, the heat-equation residual, spatial interpolation of material coefficients, boundary flux evaluation, and temporal integration are differentiable with respect to the unknown fields. The mismatch between simulated and observed surface temperatures can therefore be propagated backward through the transient heat-transfer computation to update the neural physical fields. Consequently, the governing heat-transfer model becomes part of the optimization process, and the physical fields are inferred directly through the governing physics rather than through surrogate regression from thermal appearance. Because the formulation is based on a differentiable finite-element solver, it operates naturally on irregular three-dimensional geometry, avoiding the restriction of grid-based local finite-difference stencils on regular or voxelized domains.

ThermoField can further accommodate thermal sequences acquired under different external conditions. The underlying physical fields of an object should remain consistent even when the external conditions and observed thermal trajectories change. Cooling and warming sequences, for example, probe different directions of heat flow and different balances between internal diffusion and surface exchange. Jointly optimizing such complementary observations  provides additional constraints that improve parameter identifiability. More broadly, by integrating reconstructed geometry, surface thermal observations, neural representations of thermophysical fields, and differentiable heat-transfer dynamics within a unified inverse framework, ThermoField establishes a general framework for thermophysical inference in complex three-dimensional scenes. This enables spatially resolved estimation of thermophysical properties on geometrically complex objects and support predictive heat-transfer simulation beyond the observations used for estimation.

\section{Results}
\label{sec:result}
The central hypothesis underlying ThermoField is that thermophysical parameters, such as thermal diffusivity $\alpha$ and convective heat-transfer coefficients $\beta$, can be constrained when  time-resolved thermal observations are coupled with reconstructed geometry, governing heat-transfer physics, and temporal consistency. If this hypothesis holds, the recovered fields should not only reproduce the observed thermal sequence, but also remain predictive under previously unseen thermal conditions. 
We test this hypothesis through a sequence of experiments that progressively evaluate (i) the accuracy and spatial coherence of the recovered thermophysical parameters, (ii) their predictive generalization across unseen thermal conditions, and (iii) the robustness and identifiability of the inverse optimization.

To support controlled evaluation, we construct a synthetic dataset comprising three simple geometries, and three complex objects. 
The scenes provide ground-truth geometry, material properties, and thermal boundary conditions for quantitative evaluation. 
The synthetic dataset is divided into simple and complex geometry categories, both simulated in ANSYS~\cite{ansys} to generate transient heat-transfer sequences under controlled boundary conditions.
Each object is assigned a material with distinct thermophysical and surface properties, including density, heat capacity, thermal conductivity, convective heat-transfer coefficient, and surface emissivity. Across the evaluated objects, these properties span a broad range, with thermal conductivity varying from approximately 0.2 to 149~W/(m$\cdot$K), convective coefficients from 3 to 5~W/(m$^2\cdot$K), and surface emissivity from 0.20 to 0.92, leading to substantially different thermal responses.

For each object, we consider multiple transient thermal processes that expose the material to different modes of heat transfer: (i) 30~s of external heating followed by 30~s of natural cooling, (ii) 60~s of cooling under a lower ambient temperature, and (iii) 60~s of warming under a higher ambient temperature. 
The scene configurations span ambient temperatures from $-15^\circ$C to $100^\circ$C, and applied heating powers ranging from 0 to 150~W. Training and held-out testing are separated at the scene-configuration level: training uses one cooling sequence and one heating sequence, whereas testing uses a warming sequence and a separate heating sequence with a different heating power. This split evaluates whether the recovered thermophysical properties remain predictive under unseen thermal conditions.
All simulated thermal sequences are recorded at 10~FPS, with a simulation time step set to $\Delta t = 0.1~\mathrm{s}$. Additional material properties and simulation settings are provided in Supplementary Appendix.~\ref{app:data}.

\subsection{Inverse recovery of thermophysical parameters}
A fundamental question is whether thermophysical properties can be reliably recovered from time-resolved thermal observations. We therefore evaluate ThermoField on synthetic objects with known material properties, which provide ground-truth references for quantitative assessment. Because each object is assigned a single material, the ground-truth thermal diffusivity is spatially uniform. Ideally, the recovered diffusivity field should therefore be both centered near the reference value and spatially coherent across the object surface, rather than fragmented by arbitrary local fluctuations. During optimization, the recovered distribution is expected to shift toward the ground-truth value and become more concentrated as the inferred field becomes increasingly consistent with the observed thermal evolution.

This evaluation is particularly important because ThermoField represents thermophysical properties as spatially varying fields rather than as a single global constant. We therefore assess recovery from two complementary perspectives: numerical accuracy and spatial coherence. 
Numerical accuracy is summarized by two related metrics computed for the recovered diffusivity field: the mean pointwise relative error over the evaluated surface, and the recovered/reference ratio computed from the spatial median of the recovered field with respect to the ground-truth diffusivity $\alpha_{\mathrm{GT}}$. Spatial coherence is quantified by the normalized 5--95\% interval width of the recovered field over the reconstructed surface, which measures how concentrated the inferred field remains while enabling comparison across materials with different diffusivity scales.

\begin{table*}[t]
\centering
\caption{Recovery accuracy and spatial coherence of the recovered thermophysical fields. Except for Cylinder--Cooling, all rows report thermal diffusivity with reference value $\alpha_{\mathrm{GT}}=k/(\rho c_p)$. Cylinder--Cooling instead reports the convection coefficient scaled as $\beta_{\mathrm{GT}}=h/(\rho c_p)$. Relative error denotes the mean pointwise relative error over the evaluated surface. Rec./ref. is the ratio between the spatial median of the recovered field and its reference value. Norm. width is the normalized 5--95\% interval width of the recovered field.}
\label{tab:acc}
\small
\setlength{\tabcolsep}{3.5pt}
\begin{tabular}{@{}lllcccc@{}}
\toprule
Scene & Material & Process & Target (GT)$^\dagger$ & Rel. error & Rec./ref. & Norm. width \\
\midrule
\multirow{2}{*}{Cube}
  & \multirow{2}{*}{Wood} & Cooling & Diff. (0.287) & 22.5\% & 0.778 & 33.1\% \\
  &  & Heating & Diff. (0.287) & 20.4\% & 0.842 & 65.6\% \\[1mm]
\multirow{2}{*}{Sphere}
  & \multirow{2}{*}{Copper} & Cooling & Diff. (116.306) & 42.8\% & 0.573 & 8.3\% \\
  &  & Heating & Diff. (116.306) & 14.0\% & 0.953 & 55.7\% \\[1mm]
\multirow{2}{*}{Cylinder}
  & \multirow{2}{*}{Stainless steel} & Cooling & Conv. (1.344) & 11.0\% & 1.056 & 42.8\% \\
  &  & Heating & Diff. (4.059) & 14.5\% & 1.015 & 63.1\% \\[1mm]
\multirow{2}{*}{Bunny}
  & \multirow{2}{*}{ABS} & Cooling & Diff. (0.135) & 20.7\% & 0.984 & 112.5\% \\
  &  & Heating & Diff. (0.135) & 42.6\% & 1.147 & 98.9\% \\[1mm]
\multirow{2}{*}{Bear}
  & \multirow{2}{*}{Glass} & Cooling & Diff. (0.453) & 11.9\% & 1.008 & 54.4\% \\
  &  & Heating & Diff. (0.453) & 83.1\% & 1.153 & 307.8\% \\[1mm]
\multirow{2}{*}{Car}
  & \multirow{2}{*}{Aluminum} & Cooling & Diff. (61.318) & 24.2\% & 0.889 & 84.9\% \\
  &  & Heating & Diff. (61.318) & 14.7\% & 0.938 & 54.8\% \\
\bottomrule
\end{tabular}

\vspace{2pt}
\raggedright
\footnotesize
$^\dagger$ GT values are reported in units of $10^{-6}$ of the corresponding physical quantity.
\end{table*}

Across the synthetic scenes, ThermoField recovered the target thermophysical fields with scene-dependent accuracy, reflecting differences in which physical parameters dominate the observed thermal response (Table~\ref{tab:acc}). For geometrically simple objects with nearly uniform curvature and material distribution, especially the sphere and cylinder, passive cooling does not strongly constrain thermal diffusivity. In these cases, the cooling dynamics are governed primarily by heat exchange with the environment, whereas diffusivity contributes less distinctive structure to the observable temperature evolution. This effect is most evident in Sphere--Cooling, which shows a large relative error (42.8\%) despite a narrow normalized width (8.3\%), indicating that the recovered field is spatially coherent but centered around an incorrect value. Consistently, in Cylinder--Cooling, where the estimated target is instead the convection coefficient, recovery is substantially more accurate (11.0\% relative error and a recovered/reference ratio of 1.056), supporting the interpretation that convection, rather than diffusivity, is the dominant parameter in this cooling scenario.

The heating cases show a different trend. For low-diffusivity non-metallic objects such as the bunny and bear, heat tends to accumulate locally rather than spread rapidly across the surface, making the inferred diffusivity field more sensitive to local boundary-condition mismatch and geometric reconstruction error. As a result, heating leads to substantially larger diffusivity errors in these scenes, with relative errors increasing to 42.6\% for Bunny--Heating and 83.1\% for Bear--Heating. By contrast, metallic scenes remain comparatively stable under heating, as seen in Sphere--Heating and Car--Heating, where stronger thermal excitation provides more informative constraints while rapid internal conduction helps maintain a more globally consistent thermal response.

A further trend is that complex geometries can, in some cases, make the central diffusivity estimate easier to recover because irregular shape and self-occlusion generate richer transient temperature patterns than simple canonical shapes. This behavior is reflected in the relatively accurate median-centered estimates obtained for Bunny--Cooling, Bear--Cooling and Car--Heating. 
However, accurate recovery of the central value does not necessarily imply a spatially compact field. In complex objects, geometric irregularity, visibility variation, and boundary-condition approximation can be absorbed into the inferred field, producing broader distributions and long-tailed behavior. This effect is particularly apparent in Bunny--Cooling, where the recovered/reference ratio remains close to unity (0.984) but the normalized width reaches 112.5\%, and in Bear--Heating, where the width expands to 307.8\%, indicating substantial local compensation even when the recovered field retains a physically interpretable global trend.

\subsection{Generalization across thermal conditions}

\begin{figure*}[t]
\centering
\includegraphics[width=\textwidth]{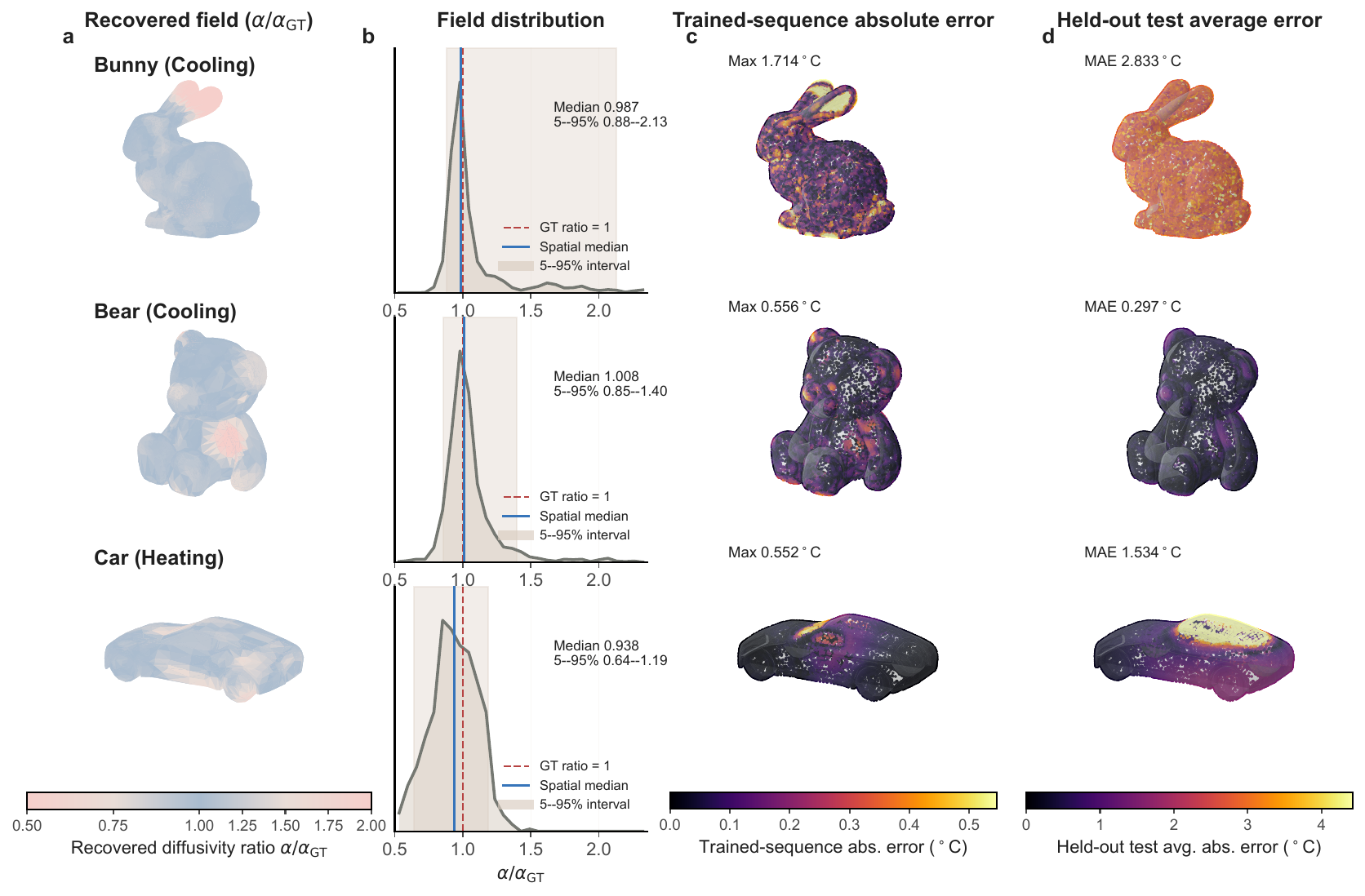}
\caption{\textbf{Representative thermophysical fields and transfer behavior across thermal conditions.}
Three representative training scenes are shown: Bunny--Cooling, Bear--Cooling, and Car--Heating. 
\textbf{a}, Recovered thermophysical field on the reconstructed geometry, shown as the recovered/reference ratio. 
\textbf{b}, Spatial distribution of the recovered field over the evaluated surface, summarized by its median and 5--95\% interval. 
\textbf{c}, Absolute temperature error on the training sequence, evaluated at the final frame for cooling scenes and at the end of the heating stage for the heating scene. 
\textbf{d}, Time-averaged absolute temperature error under the held-out test condition. 
Together, these examples show that a recovered field can remain physically interpretable and predictive across thermal conditions, while also revealing scene-dependent differences in spatial concentration and transfer accuracy.}
\label{fig:transfer}
\end{figure*}

Recovering the correct thermophysical field is only the first requirement. An equally important question is whether the recovered field represents an intrinsic material property that generalizes across thermal conditions, rather than a latent variable that merely reproduces the training sequence. We therefore evaluate cross-condition generalization by testing whether thermophysical parameters recovered from one thermal process can predict temperature evolution under held-out thermal conditions without further optimization. Representative examples of the recovered fields and their predictive behavior are shown in Fig.~\ref{fig:transfer}.

For each scene, we recover either a spatially varying thermophysical field or its spatial median from a training sequence, and then use it directly in the forward heat-transfer model under held-out test conditions. Prediction accuracy is quantified by the mean absolute error (MAE), root-mean-square error (RMSE), and maximum error over the observed temperature fields. For most scenes, the warming and heating tests follow the held-out configurations listed in Table~A2. For Cylinder--Cooling, the transferred parameter is the recovered convection field, while diffusivity is fixed to its ground-truth value in the heating test.

Across the evaluated scenes, transfer performance depends strongly on whether the parameter constrained during training remains the dominant factor under the held-out test condition (Table~\ref{tab:transfer}). When this alignment holds, the recovered spatial field remains predictive and can outperform a uniform median baseline. This behavior is most evident in Cylinder--Cooling under held-out warming, where transferring the recovered convection field yields very low error (MAE $0.038~\si{\celsius}$), whereas replacing it with a global median substantially degrades performance (MAE $11.446~\si{\celsius}$). More moderate but consistent gains from the spatial field are also observed for Cube--Cooling, Cylinder--Heating, and Bear--Cooling, indicating that the recovered non-uniform structure can encode transferable thermophysical information when the testing scenario is governed by the same underlying mechanism as the training process.

By contrast, transfer becomes less reliable when the held-out condition is governed by a different balance of physical effects, or when part of the recovered spatial variation acts as condition-specific compensation rather than intrinsic material structure. This is clearest in the heating test for Cylinder--Cooling: although the convection field recovered from passive cooling transfers well to warming, it fails under active heating, where the thermal response is determined by the coupled effects of heat input, diffusion, and surface losses rather than by boundary exchange alone. 
A similar pattern appears in the car scene, where the spatial field offers only marginal benefit under warming but performs worse than the median baseline under held-out heating, especially for Car--Heating. 

\begin{table*}[t]
\centering
\caption{Prediction accuracy under cross-condition testing. MAE, RMSE, and maximum error are reported in $^\circ$C. ``Spatial'' directly reuses the recovered spatial field, whereas ``Median'' replaces it with its spatial median as a global constant. For most scenes, warming and heating conditions follow Table~A2. For Cube, the heating test is additionally evaluated at $T_a=22^\circ\mathrm{C}$ and $\dot{Q}=1~\mathrm{W}$. For Cylinder--Cooling, the transferred parameter is the recovered convective coefficient field, while the diffusivity is fixed to the ground-truth value in the heating test. Bunny is omitted because held-out test conditions are not defined in Table~A2.}
\label{tab:transfer}
\small
\setlength{\tabcolsep}{4pt}
\begin{tabular}{llcccccc}
\toprule
\multirow{2}{*}{Training scene} & \multirow{2}{*}{Rep.}
& \multicolumn{3}{c}{Test warming}
& \multicolumn{3}{c}{Test heating} \\
\cmidrule(lr){3-5} \cmidrule(lr){6-8}
& & MAE & RMSE & Max. error & MAE & RMSE & Max. error \\
\midrule
Cube--Cooling   & Spatial & 0.077 & 0.102 & 0.700 & 0.665 & 1.395 & 10.418 \\
Cube--Cooling   & Median  & 0.096 & 0.134 & 0.890 & 0.875 & 1.845 & 10.736 \\
Cube--Heating   & Spatial & 0.134 & 0.199 & 1.485 & 0.297 & 0.524 & 4.244 \\
Cube--Heating   & Median  & 0.095 & 0.147 & 0.980 & 0.453 & 0.962 & 7.714 \\
\midrule
Cylinder--Cooling & Spatial & 0.038 & 0.045 & 0.113 & 45.039 & 48.959 & 82.861 \\
Cylinder--Cooling & Median  & 11.446 & 12.957 & 21.195 & 36.342 & 39.179 & 73.912 \\
Cylinder--Heating & Spatial & 0.054 & 0.062 & 0.157 & 0.516 & 0.585 & 3.265 \\
Cylinder--Heating & Median  & 0.116 & 0.123 & 0.223 & 2.588 & 3.684 & 9.772 \\
\midrule
Bear--Cooling   & Spatial & 0.196 & 0.270 & 1.695 & 0.386 & 1.788 & 38.104 \\
Bear--Cooling   & Median  & 0.245 & 0.381 & 3.415 & 0.397 & 1.830 & 36.794 \\
Bear--Heating   & Spatial & 0.784 & 0.943 & 3.369 & 0.306 & 1.727 & 33.805 \\
Bear--Heating   & Median  & 0.805 & 0.895 & 3.299 & 0.315 & 1.850 & 35.972 \\
\midrule
Car--Cooling    & Spatial & 0.485 & 0.552 & 1.115 & 2.186 & 5.388 & 49.441 \\
Car--Cooling    & Median  & 0.523 & 0.594 & 1.220 & 0.861 & 1.256 & 6.198 \\
Car--Heating    & Spatial & 0.485 & 0.554 & 1.176 & 1.505 & 2.832 & 25.933 \\
Car--Heating    & Median  & 0.522 & 0.593 & 1.221 & 0.050 & 0.088 & 0.778 \\
\bottomrule
\end{tabular}
\end{table*}

However, successful transfer under unseen thermal conditions does not by itself imply that the recovered parameters are uniquely or stably identifiable. An equally important question is whether repeated inverse optimizations converge to similar thermophysical fields, or whether multiple distinct solutions can explain the observations with comparable accuracy.

\subsection{Robustness and identifiability under repeated inverse optimization}

A remaining question is whether the recovered thermophysical fields are robust to the non-convex optimization underlying the inverse problem. If the recovered parameters correspond to intrinsic physical properties, repeated optimizations starting from different initializations should converge to similar thermophysical fields rather than substantially different solutions. We therefore evaluate the robustness of ThermoField under repeated inverse optimization. For each scenario, we repeat the optimization 16 times using four initialization settings and four random seeds. This experimental design separates two sources of variability: stochastic optimization effects arising from different random seeds and systematic variability introduced by different initializations.

\begin{figure}[h]
\centering
\includegraphics[width=1.0\textwidth]{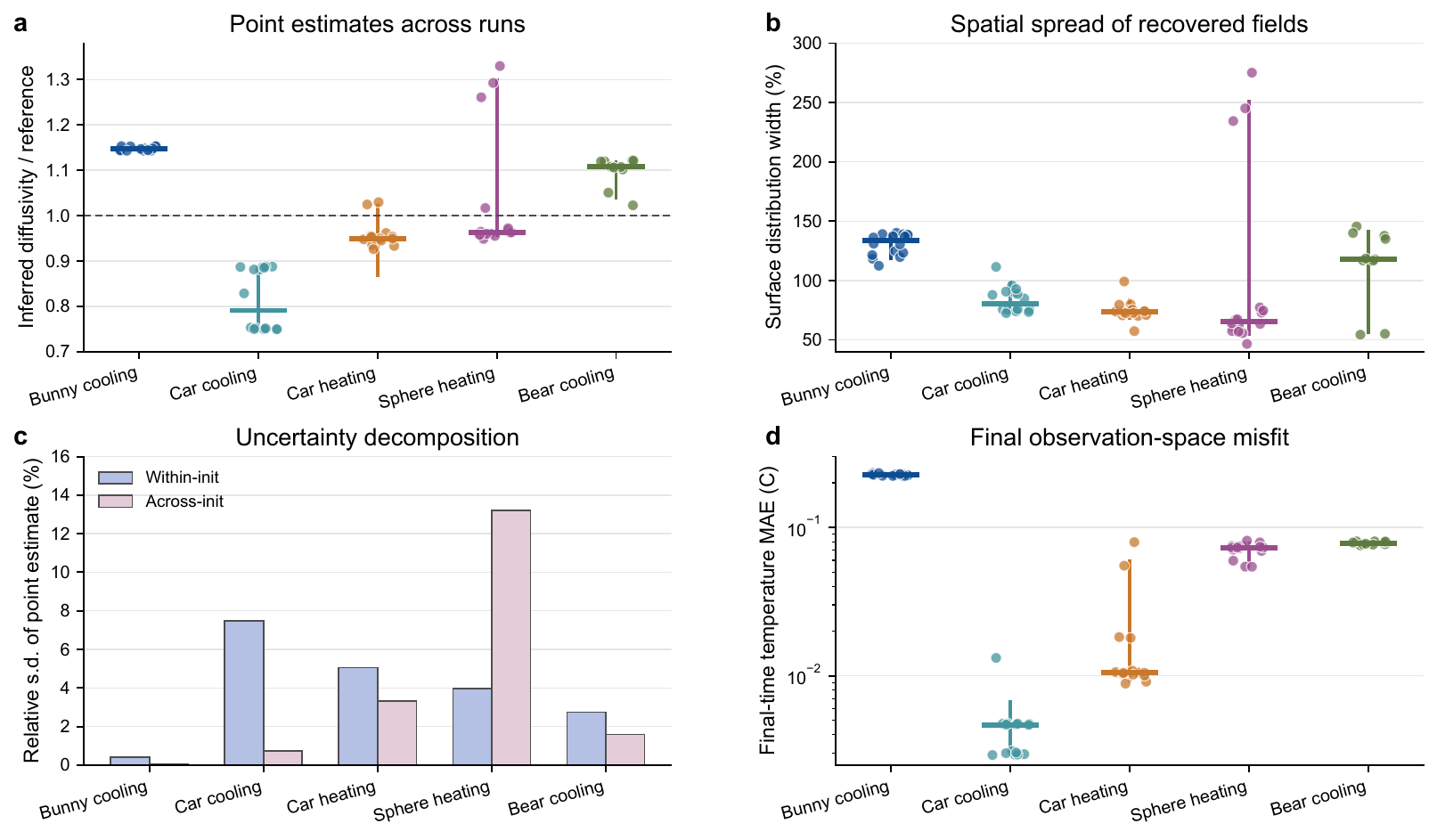}
\caption{\textbf{Robustness and identifiability of recovered thermophysical fields.}
\textbf{a}, Point estimates across repeated runs, reported as recovered/reference ratios.
\textbf{b}, Spatial spread of the recovered diffusivity field, quantified by the normalized 5--95\% interval width over the reconstructed surface.
\textbf{c}, Decomposition of variability into within-initialization and across-initialization components.
\textbf{d}, Final observation-space temperature error, measured by the final-time temperature MAE.}
\label{fig}
\end{figure}

Across the repeated-run ensemble, the recovered point estimates are generally consistent within each scenario, although not always unbiased with respect to the reference material value (Fig.~\ref{fig}a).
The median inferred diffusivity is 1.147 for bunny cooling, 0.791 for car cooling, 0.949 for car heating,0.963 for sphere heating, and 1.107 for bear cooling.  These results indicate that ThermoField is generally robust across runs, while also showing that repeatability alone does not guarantee correctness: for example, car cooling remains systematically below the reference value, whereas sphere heating stays near the reference in most runs but exhibits occasional larger deviations.

Parameter robustness should also be reflected in the spatial consistency of the recovered fields rather than only their scalar values. The median normalized 5--95\% widths are 133.7\% for bunny cooling, 80.3\% for car cooling, 73.9\% for car heating,  65.5\% for sphere heating, and 118.2\% for bear cooling (Fig.~\ref{fig}b). Consistent with the parameter recovery and generalization experiments, heating-based estimates are generally more spatially concentrated, suggesting that stronger thermal excitation improves the conditioning of spatial field recovery. Cooling-only scenarios show broader distributions, indicating that passive temperature evolution provides weaker constraints on local material-property variations.

To distinguish optimization variability from inherent ambiguity in the inverse problem, we decompose the run-to-run variability into within-initialization and across-initialization components (Fig.~\ref{fig}c). Across-initialization variability is most pronounced for sphere heating, indicating stronger dependence on the optimization basin. Bunny cooling and car cooling are more affected by within-initialization dispersion, whereas car heating shows a more balanced contribution from initialization and seed variability. Bear cooling also shows non-negligible variability, consistent with a weaker degree of identifiability under passive cooling.

Despite this parameter variability, the observation-space temperature error remains small in most scenarios (Fig.~\ref{fig}d). The median final-time MAEs are 0.227~\si{\celsius} for bunny cooling, 0.0047~\si{\celsius} for car cooling, 0.0105~\si{\celsius} for car heating, 0.0731~\si{\celsius} for sphere heating, and 0.0783~\si{\celsius} for bear cooling. This mismatch between observation-space accuracy and parameter variability illustrates a fundamental property of thermophysical inverse problems: several nearby, or occasionally distinct, parameter configurations can reproduce the observed temperature evolution with comparable accuracy. Consequently, low observation-space reconstruction error should not be interpreted as evidence of unique parameter recovery. Assessing parameter identifiability requires analyzing the distribution of recovered thermophysical fields across repeated optimizations.

These experiments provide two complementary insights. First, \textit{ThermoField} recovers physically plausible thermophysical fields across repeated optimizations despite the non-convex inverse formulation. Second, the variability across repeated runs provides a practical measure of the conditioning and identifiability of the underlying inverse problem. Rather than returning a single optimized solution, ThermoField characterizes a distribution of physically plausible thermophysical fields, enabling uncertainty-aware interpretation of inverse thermophysical estimation.

\section{Discussion}

ThermoField demonstrates that thermophysical properties can be recovered directly from time-resolved thermal observations of complex three-dimensional objects by combining reconstructed geometry with differentiable heat-transfer physics. Rather than treating thermal images as appearance to be reconstructed, the framework formulates thermophysical inference on reconstructed scenes, enabling material properties to be estimated within a physically constrained inverse problem. A key aspect of the framework is its use of a signed-distance-function-based surface representation. This representation is well aligned with thermal sensing because thermal cameras observe radiometric emission from object surfaces rather than volumetric temperature fields directly. By reconstructing the surface geometry on which thermal measurements are defined, the framework establishes the computational domain on which thermophysical fields are defined and the transient heat-transfer model is solved, enabling simulated temperature evolution to be compared directly with the observed thermal sequence.

A central finding of this work is that thermophysical properties are more naturally represented as spatially varying fields than as single global parameters. This differs from approaches that infer a single global value for a material or object. In practice, geometry reconstruction, thermal calibration, and differentiable finite-element simulation inevitably introduce numerical errors. Under such conditions, a single optimized scalar can be sensitive to local errors, discretization artifacts, or extreme predictions. Representing thermophysical properties as spatial fields provides substantially richer information than a single optimized scalar. The recovered field reveals whether the inferred material properties remain spatially coherent or whether the optimization compensates for modeling inaccuracies through localized parameter variations. Consequently, the inferred field serves not only as an estimate of the underlying material properties but also as a diagnostic of parameter identifiability and the conditioning of the inverse problem.

An important consequence of recovering thermophysical fields rather than thermal appearance is that the inferred properties can  be directly coupled to forward heat-transfer simulations. Once estimated, the field parameters are not restricted to reproducing the observed thermal sequence, but can be used to predict temperature evolution under new environmental or boundary conditions. This distinguishes ThermoField from thermal scene reconstruction methods, whose primary objective is reproducing observed thermal sequences rather than recovering transferable physical properties. While such methods can produce spatially consistent thermal maps or time-varying thermal representations, they are often optimized to fit the captured observations and do not necessarily provide parameters that support counterfactual thermal prediction. By contrast, the proposed framework connects reconstruction with physical simulation, enabling the recovered fields to support prediction beyond the acquisition conditions.

The principal limitation of the current framework is not the differentiable formulation itself but the amount of physical information available in the observed thermal sequence. Parameter identifiability fundamentally depends on whether the measured temperature evolution contains sufficient information to distinguish different thermophysical configurations. Metallic objects, or materials with high thermal diffusivity, tend to equilibrate temperature rapidly and may exhibit weak spatial temperature gradients under mild excitation, making diffusivity more difficult to infer from surface observations alone. Non-metallic materials often produce stronger surface-temperature transients and more persistent gradients between heated and unheated regions, which provide more informative signals for estimating diffusivity. Similarly, passive natural cooling or warming generally provides weaker constraints than external heating, because the induced temperature changes are smaller and less spatially structured. Spherical objects represent a particularly challenging case under approximately uniform boundary conditions: their symmetry produces limited surface-temperature variation, leaving few spatial gradients from which diffusivity can be identified. These observations emphasize that the primary challenge in thermophysical scene inference is parameter identifiability rather than optimization. Even with a differentiable forward model, reliable recovery requires sufficiently informative thermal excitation, observable temperature gradients, and accurate knowledge of geometry and boundary conditions.

Several extensions could further improve the scope of the framework. Future work could address these limitations by incorporating richer excitation protocols, active heating patterns, multi-view or multi-spectral thermal measurements, and stronger priors on material classes or boundary conditions. Extending the framework to jointly estimate additional quantities, such as surface emissivity, convective heat-transfer coefficients, and spatially varying boundary interactions, would further improve its applicability to real-world scenes. More broadly, this work suggests that thermal vision can evolve from reconstructing temperature appearance toward recovering physically meaningful scene representations. Such representations provide a foundation for thermophysical scene inference, predictive simulation, and digital twins that remain consistent across changing environmental conditions. Extending this paradigm beyond thermal diffusion to broader classes of physical processes could enable vision systems that recover latent physical properties rather than merely reproducing observations.

\section{Methodology}
\label{sec:method}

\textbf{Overview } 
ThermoField formulates thermophysical inference as a differentiable inverse problem that combines reconstructed three-dimensional geometry, time-resolved thermal observations, and governing heat-transfer physics to recover spatially varying thermophysical fields. Unlike conventional inverse heat-transfer methods, which typically operate on prescribed simplfied computational domains, ThermoField performs thermophysical inference directly on reconstructed three-dimensional scenes, enabling material properties to be recovered from thermal observations of geometrically complex objects.
An overview of the proposed method is shown in Fig.~\ref{fig:architecture}.
Additional implementation details, including the SDF reconstruction objective, finite-element discretization, and optimization settings, are provided in the Appendix.~\ref{app:method}.

The framework consists of four main components. First, the object geometry is reconstructed from multi-view RGB images and represented as a signed distance field (SDF), providing a continuous representation of the object surface. Second, the reconstructed geometry is recovered at metric scale and serves as the computational domain for thermophysical inference. The observed thermal sequence is registered onto the reconstructed surface, establishing correspondence between the measured temperature evolution and the underlying geometry. 
Third, spatially varying thermophysical quantities are represented as neural fields defined on sparse surface control points and interpolated onto the reconstructed mesh. Finally, these fields are embedded into a differentiable transient heat-transfer simulation, and optimized by minimizing the discrepancy between the simulated and observed surface-temperature evolution.

The following sections describe the individual components of this framework. We first introduce the  geometric reconstruction and heat transfer models that establish the foundation for thermophysical inference.
Because the observed thermal dynamics depend jointly on object geometry and  material properties, accurate geometric reconstruction is a prerequisite for reliable recovery of thermophysical fields.

\subsection{SDF-based geometry reconstruction}
To recover the object geometry from multi-view RGB observations, we adopt an implicit neural representation based on signed distance functions (SDFs). In this representation, the scene is reconstructed by a signed distance function $f(\mathbf{x})$, which maps a spatial position $\mathbf{x}\in\mathbb{R}^{3}$ to its signed distance from the object surface, and a color function that supports differentiable volume rendering. The object surface $\mathcal{S}$ is represented as the zero-level set of the SDF:
\begin{equation}
\mathcal{S}=\{\mathbf{x}\in\mathbb{R}^{3}\mid f(\mathbf{x})=0\},
\label{eq:sdf}
\end{equation}

Following NeuS~\cite{wang2021neus}, the geometry and appearance networks are jointly optimized by minimizing the discrepancy between rendered and observed RGB images from multiple viewpoints, together with standard SDF regularization. After optimization, the recovered surface is extracted from the learned zero level set and converted into a tetrahedral mesh for physical simulation.  Unlike conventional neural rendering approaches, where the reconstructed geometry is primarily used for novel-view synthesis, ThermoField uses the reconstructed surface to define the computational domain for thermophysical inversion.

In contrast to visual reconstruction alone, thermophysical inference requires geometry at the correct physical scale, because the spatial derivatives appearing in the heat-transfer equations depend directly on the object's dimensions. We therefore recover the metric scale of the reconstructed geometry using external scale cues derived from metric depth estimation. The resulting metrically aligned surface and tetrahedral mesh define the domain on which thermophysical fields are represented and transient heat-transfer simulation is performed. Because geometry reconstruction is not the main methodological contribution of this work, additional details of the SDF representation, volume-rendering formulation, and optimization procedure are provided in the Supplementary Appendix~\ref{app:sdf}.

The reconstructed geometry establishes the spatial domain required for thermophysical inference. We next introduce the governing equations of transient heat transfer that relate the reconstructed geometry and thermophysical properties to the observed temperature evolution.

\subsection{Heat transport and spatial thermophysical fields}

Temperature evolution inside the reconstructed object is governed by transient heat conduction,
\begin{equation}
\rho c_{p}\frac{\partial T}{\partial t}
=
\nabla \cdot (k \nabla T)
+
\dot{q},
\label{eq:heat}
\end{equation}
where $T=T(\mathbf{x},t)$ denotes the temperature field, $\rho$ is the material density, $c_p$ is the specific heat capacity, $k$ is the thermal conductivity, and $\dot{q}$ is the volumetric heat generation rate. The physical quantities used in the model are summarized in Table~\ref{tab:units_heat_app}.

In practical settings, heat exchange at object surfaces is additionally affected by convection and thermal radiation. Consequently, the heat transport model must satisfy   boundary conditions at the object surface that account for these effects:
\begin{equation}
-k \nabla T \cdot \mathbf{n} = h(T_s - T_{\infty}) + \varepsilon \sigma \left(T_s^{4} - T_{\mathrm{env}}^{4}\right),
\label{eq:boundary}
\end{equation}
where $\mathbf{n}$ denotes the outward surface normal, $h$ is the convective heat-transfer coefficient, $\varepsilon$ is the surface emissivity, $\sigma$ is the Stefan--Boltzmann constant, $T_{\infty}$ is the ambient fluid temperature, $T_{\mathrm{env}}$ is the effective surrounding radiation temperature, and $q_b$ denotes externally applied boundary heating when present.

In the diffusivity-estimation setting considered in this work, it is convenient to express the governing material parameter in terms of thermal diffusivity,
\begin{equation}
\alpha = \frac{k}{\rho c_p},
\end{equation}
and, analogously, to express the surface-exchange terms in scaled form as
\begin{equation}
\beta = \frac{h}{\rho c_p},
\qquad
\gamma = \frac{\varepsilon}{\rho c_p},
\end{equation}
which are the quantities directly represented by the neural physical fields in the inverse formulation.


To make this formulation operational, ThermoField represents thermophysical quantities as explicit spatial fields on the reconstructed object. Let $\phi(\mathbf{x})$ denote a generic thermophysical field, such as $\alpha(\mathbf{x})$, $\beta(\mathbf{x})$, or $\gamma(\mathbf{x})$. The unknown fields are parameterized on sparse surface control points sampled from the reconstructed geometry. Their coordinates are encoded using positional features and mapped to field values by a multilayer perceptron. The raw network outputs are transformed through a sigmoid function and bounded within prescribed physical ranges.

The predicted control-point values are then interpolated to the boundary vertices of the reconstructed mesh and further extended into the tetrahedral cells used by the forward solver. This formulates spatially resolved thermophysical fields defined consistently over the reconstructed geometry. Rather than reducing the object to a single global material constant, this formulation can represent local variations associated with heterogeneous materials, surface treatments, geometric irregularities, or region-dependent thermal responses, while also exposing the spatial distribution of the recovered field itself.

\subsection{Differentiable inverse thermophysical optimization}

Given thermophysical fields and external thermal conditions, ThermoField simulates transient temperature evolution on the reconstructed tetrahedral mesh by solving Eqs.~(\ref{eq:heat})--(\ref{eq:boundary}) with a differentiable finite-element solver implemented in JAX-FEM. In this formulation, the reconstructed geometry defines the simulation domain, the thermophysical fields determine the cell-wise material properties and boundary exchange coefficients, and the forward solve produces the full temperature trajectory in space and time. The radiative term is linearized around the previous time step when evaluating the boundary flux, which improves numerical stability while preserving differentiability of the overall transient solve.

Because the solver is differentiable, the residual assembly, boundary-flux evaluation, implicit time stepping, and solution operator are embedded in the automatic-differentiation pipeline. The mismatch between simulated and observed surface temperatures can therefore be propagated backward through the full transient heat-transfer computation to update the neural thermophysical fields. In this way, the governing heat-transfer model becomes part of the inverse optimization process, and the physical fields are inferred directly through the governing physics rather than through surrogate regression from thermal appearance.

Given observed surface temperatures $T_s^{\mathrm{obs}}(\mathbf{x}_i,t)$ at sampled surface locations $\mathbf{x}_i$ and times $t$, we optimize the field parameters $\Theta$ by minimizing the discrepancy between observed and simulated temperatures,
\begin{equation}
\mathcal{L}_{\mathrm{data}}
=
\frac{1}{N}
\sum_{i,t}
\left\|
\hat{T}_s(\mathbf{x}_i,t;\Theta) - T_s^{\mathrm{obs}}(\mathbf{x}_i,t)
\right\|_2^2,
\end{equation}
where $\hat{T}_s(\mathbf{x}_i,t;\Theta)$ denotes the simulated surface temperature trajectory. In practice, the simulated temperature field is computed at the boundary vertices of the tetrahedral mesh and then interpolated back to the observed surface sample points using precomputed geometric correspondences. Supervision is therefore applied directly in the observation space.

\noindent\textbf{Linear and quadratic simulation stages.}
To balance numerical stability and simulation accuracy, ThermoField adopts a staged finite-element discretization strategy. The inverse optimization is first performed on a linear tetrahedral mesh (TET4) and then refined on a quadratic tetrahedral mesh (TET10). The linear stage provides a cheaper and more stable initialization, helping the optimization first recover a plausible coarse thermophysical field. The quadratic stage then improves interpolation accuracy and more faithfully resolves heat transport on complex geometry. This coarse-to-fine strategy improves practical convergence while retaining the higher accuracy of quadratic finite elements in the final recovered solution.

\noindent\textbf{Field smoothness.}
Because thermophysical quantities are represented as spatial fields, the inverse problem contains substantially more freedom than scalar parameter estimation. Although this flexibility is useful for modeling non-uniform thermal behavior, it also makes the optimization prone to non-physical local oscillations. If only the data term is used, the model can match the observed thermal sequence by introducing fragmented local excursions in the recovered field, producing broad spatial distributions and large 5--95\% interval widths.

To suppress such behavior, we impose a smoothness regularization term over neighboring surface control points,
\begin{equation}
\mathcal{L}_{\mathrm{smooth}}
=
\sum_{\phi \in \mathcal{F}}
\frac{1}{|\mathcal{N}|}
\sum_{(p_i,p_j)\in\mathcal{N}}
\left(
\phi(p_i)-\phi(p_j)
\right)^2,
\end{equation}
where $\mathcal{F}$ denotes the set of trainable thermophysical fields and $\mathcal{N}$ denotes neighboring control-point pairs on the reconstructed surface. This term encourages nearby control points to take similar values and promotes spatial coherence of the recovered fields.

The final optimization objective is
\begin{equation}
\mathcal{L}
=
\mathcal{L}_{\mathrm{data}}
+
\lambda_{\mathrm{smooth}}\mathcal{L}_{\mathrm{smooth}}
+
\lambda_{\mathrm{reg}}\mathcal{L}_{\mathrm{reg}},
\end{equation}
where $\mathcal{L}_{\mathrm{reg}}$ denotes standard parameter regularization on the neural field parameters. Empirically, weakening or removing the smoothness term leads to substantially broader field distributions, indicating that the optimizer increasingly explains the observations through local compensation rather than through a physically coherent thermophysical field.

Overall, ThermoField integrates metric geometry reconstruction, spatial neural thermophysical fields, differentiable heat-transfer simulation, staged finite-element refinement, and smoothness-regularized inverse optimization into a unified framework for thermophysical inference from time-resolved thermal observations.

\section*{Acknowledgment}
This research was supported by Innosuisse - Swiss Innovation Agency under Grant \textbf{No. 105.237.1 IP-ICT}, titled \textit{Insulated: Integrated Solution for Lean and Abridged Thermal Evaluation with Digital Twins}, a collaborative project between EPFL and Schindler AG.

\bibliography{sn-bibliography}

@String(CVPR  = {IEEE Conf. Comput. Vis. Pattern Recog.})

@String(ECCV  = {Eur. Conf. Comput. Vis.})

@String(ICLR  = {Int. Conf. Learn. Represent.})

@String(CVPR  = {CVPR})

@String(ECCV  = {ECCV})

@String(ICLR  = {ICLR})

@inbook{ch4,
publisher = {John Wiley I\& Sons, Ltd},
isbn = {9783527693306},
title = {Some Basic Concepts in Heat Transfer},
booktitle = {Infrared Thermal Imaging},
chapter = {4},
pages = {351-392},
doi = {https://doi.org/10.1002/9783527693306.ch4},
url = {https://onlinelibrary.wiley.com/doi/abs/10.1002/9783527693306.ch4},
eprint = {https://onlinelibrary.wiley.com/doi/pdf/10.1002/9783527693306.ch4},
year = {2017}
}

@ARTICLE{Chen24physical,
        author={Chen, Jiaqi and Frey, Jonas and Zhou, Ruyi and Miki, Takahiro and Martius, Georg and Hutter, Marco},
        journal={IEEE Robotics and Automation Letters}, 
        title={Identifying Terrain Physical Parameters From Vision - Towards Physical-Parameter-Aware Locomotion and Navigation}, 
        year={2024},
        volume={9},
        number={11},
        pages={9279-9286},
        doi={10.1109/LRA.2024.3455788}}

@InProceedings{10.1007/978-3-031-99565-1_8,
author="Cardoso, Ricardo Pedreiras
and Moreno, Plinio",
editor="Gon{\c{c}}alves, Nuno
and Oliveira, H{\'e}lder P.
and S{\'a}nchez, Joan Andreu",
title="Estimating Object Physical Properties from RGB-D Vision and Depth Robot Sensors Using Deep Learning",
booktitle="Pattern Recognition and Image Analysis",
year="2026",
publisher="Springer Nature Switzerland",
address="Cham",
pages="97--110",
}

@inproceedings{PhysAavatar24,
    title={PhysAvatar: Learning the Physics of Dressed 3D Avatars from Visual Observations},
    author={Yang Zheng and Qingqing Zhao and Guandao Yang and Wang Yifan and Donglai Xiang and Florian Dubost and Dmitry Lagun and Thabo Beeler and Federico Tombari and Leonidas Guibas and Gordon Wetzstein},
    journal={European Conference on Computer Vision (ECCV)},
    year={2024}
}

@article{zhao2024efficient,
  title={Efficient Physics Simulation for 3D Scenes via MLLM-Guided Gaussian Splatting},
  author={Zhao, Haoyu and Wang, Hao and Zhao, Xingyue and Fei, Hao and Wang, Hongqiu and Long, Chengjiang and Zou, Hua},
  journal={arXiv preprint arXiv:2411.12789},
  year={2024}
}

@inproceedings{chen2025vid2sim,
  title={Vid2Sim: Generalizable, Video-based Reconstruction of Appearance, Geometry and Physics for Mesh-free Simulation},
  author={Chen, Chuhao and Dou, Zhiyang and Wang, Chen and Huang, Yiming and Chen, Anjun and Feng, Qiao and Gu, Jiatao and Liu, Lingjie},
  booktitle={Proceedings of the Computer Vision and Pattern Recognition Conference},
  pages={26545--26555},
  year={2025}
}

@inproceedings{dashpute2023thermal,
  title={Thermal spread functions (tsf): Physics-guided material classification},
  author={Dashpute, Aniket and Saragadam, Vishwanath and Alexander, Emma and Willomitzer, Florian and Katsaggelos, Aggelos and Veeraraghavan, Ashok and Cossairt, Oliver},
  booktitle={Proceedings of the IEEE/CVF Conference on Computer Vision and Pattern Recognition},
  pages={1641--1650},
  year={2023}
}

@article{bryan1996inverse,
  title={An inverse problem in thermal imaging},
  author={Bryan, Kurt and Caudill, Jr, Lester F},
  journal={SIAM journal on Applied Mathematics},
  volume={56},
  number={3},
  pages={715--735},
  year={1996},
  publisher={SIAM}
}

@article{https://doi.org/10.1111/j.1467-8659.2003.00716.x,
author = {Patow, Gustavo and Pueyo, Xavier},
title = {A Survey of Inverse Rendering Problems},
journal = {Computer Graphics Forum},
volume = {22},
number = {4},
pages = {663-687},
doi = {https://doi.org/10.1111/j.1467-8659.2003.00716.x},
url = {https://onlinelibrary.wiley.com/doi/abs/10.1111/j.1467-8659.2003.00716.x},
eprint = {https://onlinelibrary.wiley.com/doi/pdf/10.1111/j.1467-8659.2003.00716.x},
year = {2003}
}

@inproceedings{wu2025pbrnerf,
      title     = {{PBR-NeRF}: Inverse Rendering with Physics-Based Neural Fields},
      author    = {Wu, Sean and Basu, Shamik and Broedermann, Tim and Van Gool, Luc and Sakaridis, Christos},
      booktitle = {Proceedings of the IEEE/CVF Conference on Computer Vision and Pattern Recognition (CVPR)},
      year      = {2025}
}

@InProceedings{humanMotionKZFM19,
  title={Learning 3D Human Dynamics from Video},
  author = {Angjoo Kanazawa and Jason Y. Zhang and Panna Felsen and Jitendra Malik},
  booktitle={Computer Vision and Pattern Recognition (CVPR)},
  year={2019}
}

@InProceedings{Li_NeISF_CVPR2024,
    author    = {Li, Chenhao and Ono, Taishi and Uemori, Takeshi and Mihara, Hajime and Gatto, Alexander and Nagahara, Hajime and Moriuchi, Yusuke},
    title     = {NeISF: Neural Incident Stokes Field for Geometry and Material Estimation},
    booktitle = {Proceedings of the IEEE/CVF Conference on Computer Vision and Pattern Recognition (CVPR)},
    month     = {June},
    year      = {2024},
    pages     = {21434-21445}
}

@inproceedings{10.1007/978-3-030-58583-9_1,
author = {Purri, Matthew and Dana, Kristin},
title = {Teaching Cameras to Feel: Estimating Tactile Physical Properties of Surfaces from Images},
year = {2020},
isbn = {978-3-030-58582-2},
publisher = {Springer-Verlag},
address = {Berlin, Heidelberg},
doi = {10.1007/978-3-030-58583-9_1},
booktitle = {Computer Vision – ECCV 2020: 16th European Conference, Glasgow, UK, August 23–28, 2020, Proceedings, Part XXVII},
pages = {1–20},
numpages = {20},
location = {Glasgow, United Kingdom}
}

@inproceedings{10.5555/3666122.3669051,
author = {Tung, Hsiao-Yu and Ding, Mingyu and Chen, Zhenfang and Bear, Daniel M. and Gan, Chuang and Tenenbaum, Joshua B. and Yamins, Daniel L. K. and Fan, Judith and Smith, Kevin A.},
title = {Physion++: evaluating physical scene understanding that requires online inference of different physical properties},
year = {2023},
publisher = {Curran Associates Inc.},
address = {Red Hook, NY, USA},
booktitle = {Proceedings of the 37th International Conference on Neural Information Processing Systems},
articleno = {2929},
numpages = {21},
location = {New Orleans, LA, USA},
series = {NIPS '23}
}

@inproceedings{chow2025physbench,
  title={PhysBench: Benchmarking and Enhancing Vision-Language Models for Physical World Understanding},
  author={Chow, Wei and Mao, Jiageng and Li, Boyi and Seita, Daniel and Guizilini, Vitor and Wang, Yue},
  booktitle={The Thirteenth International Conference on Learning Representations (ICLR)},
  year={2025},
}

@misc{zhou2026digitaltwinaiopportunities,
      title={Digital Twin AI: Opportunities and Challenges from Large Language Models to World Models}, 
      author={Rong Zhou and Dongping Chen and Zihan Jia and Yao Su and Yixin Liu and Yiwen Lu and Dongwei Shi and Yue Huang and Tianyang Xu and Yi Pan and Xinliang Li and Yohannes Abate and Qingyu Chen and Zhengzhong Tu and Yu Yang and Yu Zhang and Qingsong Wen and Gengchen Mai and Sunyang Fu and Jiachen Li and Xuyu Wang and Ziran Wang and Jing Huang and Tianming Liu and Yong Chen and Lichao Sun and Lifang He},
      year={2026},
      eprint={2601.01321},
      archivePrefix={arXiv},
      primaryClass={cs.AI},
      url={https://arxiv.org/abs/2601.01321}, 
}

@InProceedings{Li_2025_CVPR,
    author    = {Li, Wenqiao and Gu, Yao and Chen, Xintao and Xu, Xiaohao and Hu, Ming and Huang, Xiaonan and Wu, Yingna},
    title     = {Towards Visual Discrimination and Reasoning of Real-World Physical Dynamics: Physics-Grounded Anomaly Detection},
    booktitle = {Proceedings of the IEEE/CVF Conference on Computer Vision and Pattern Recognition (CVPR)},
    month     = {June},
    year      = {2025},
    pages     = {30409-30419}
}

@inproceedings{pgvlm2024,
    title={Physically Grounded Vision-Language Models for Robotic Manipulation},
    author={Jensen Gao and Bidipta Sarkar and Fei Xia and Ted Xiao and Jiajun Wu
    and Brian Ichter and Anirudha Majumdar and Dorsa Sadigh},
    booktitle={IEEE International Conference on Robotics and Automation (ICRA)},
    year={2024},
    organization={IEEE}
}

@inbook{physics_of_thermal,
author = {DeWitt, D. P. and Incropera, F. P.},
publisher = {John Wiley I\& Sons, Ltd},
isbn = {9780470172575},
title = {Physics of Thermal Radiation},
booktitle = {Theory and Practice of Radiation Thermometry},
chapter = {1},
pages = {19-89},
doi = {https://doi.org/10.1002/9780470172575.ch1},
url = {https://onlinelibrary.wiley.com/doi/abs/10.1002/9780470172575.ch1},
eprint = {https://onlinelibrary.wiley.com/doi/pdf/10.1002/9780470172575.ch1},
year = {1988}
}

@inproceedings{10.1117/12.3065037,
author = {Jean-Claude Krapez},
title = {{Multiwavelength thermometry without a priori emissivity information: from promise to disillusionment}},
volume = {13470},
booktitle = {Thermosense: Thermal Infrared Applications XLVII},
editor = {Giovanni Ferrarini and Peter Spaeth and Fernando L{\'o}pez},
organization = {International Society for Optics and Photonics},
publisher = {SPIE},
pages = {1347006},
year = {2025},
doi = {10.1117/12.3065037},
URL = {https://doi.org/10.1117/12.3065037}
}

@InProceedings{Narayanan_2026_CVPR,
        author    = {Narayanan, Sriram and Ramanagopal, Mani and Narasimhan, Srinivasa},
        title     = {Dual Band Thermal Videography: Separating Time-Varying Reflection and Emission Near Ambient Conditions},
        booktitle = {Proceedings of the IEEE/CVF Conference on Computer Vision and Pattern Recognition (CVPR)},
        month     = {June},
        year      = {2026},
        pages     = {199-208}
    }

@article{WANG2023105640,
title = {Thermal images-aware guided early fusion network for cross-illumination RGB-T salient object detection},
journal = {Engineering Applications of Artificial Intelligence},
volume = {118},
pages = {105640},
year = {2023},
issn = {0952-1976},
doi = {https://doi.org/10.1016/j.engappai.2022.105640},
url = {https://www.sciencedirect.com/science/article/pii/S0952197622006303},
author = {Han Wang and Kechen Song and Liming Huang and Hongwei Wen and Yunhui Yan}}

@ARTICLE{9108585,
  author={Sun, Yuxiang and Zuo, Weixun and Yun, Peng and Wang, Hengli and Liu, Ming},
  journal={IEEE Transactions on Automation Science and Engineering}, 
  title={FuseSeg: Semantic Segmentation of Urban Scenes Based on RGB and Thermal Data Fusion}, 
  year={2021},
  volume={18},
  number={3},
  pages={1000-1011},
  doi={10.1109/TASE.2020.2993143}}

@article{vollmer2024detecting,
  title={Detecting district heating leaks in thermal imagery: Comparison of anomaly detection methods},
  author={Vollmer, Elena and Ruck, Julian and Volk, Rebekka and Schultmann, Frank},
  journal={Automation in Construction},
  volume={168},
  pages={105709},
  year={2024},
  publisher={Elsevier}
}

@article{hassan2025thermonerf,
  title={ThermoNeRF: A multimodal Neural Radiance Field for joint RGB-thermal novel view synthesis of building facades},
  author={Hassan, Mariam and Forest, Florent and Fink, Olga and Mielle, Malcolm},
  journal={Advanced Engineering Informatics},
  volume={65},
  pages={103345},
  year={2025},
  publisher={Elsevier}
}

@inproceedings{chen2024thermal3dgs,
  title={Thermal3D-GS: Physics-induced 3D Gaussians for
Thermal Infrared Novel-view Synthesis},
  author={Chen, Qian and Shu, shihao and Bai, Xiangzhi},
  booktitle={European Conference on Computer Vision},
  year={2024}
}

@article{hu2019difftaichi,
  title={DiffTaichi: Differentiable Programming for Physical Simulation},
  author={Hu, Yuanming and Anderson, Luke and Li, Tzu-Mao and Sun, Qi and Carr, Nathan and Ragan-Kelley, Jonathan and Durand, Fr{\'e}do},
  journal={ICLR},
  year={2020}
}

@inproceedings{NIPS2015_d09bf415,
 author = {Wu, Jiajun and Yildirim, Ilker and Lim, Joseph J and Freeman, Bill and Tenenbaum, Josh},
 booktitle = {Advances in Neural Information Processing Systems},
 editor = {C. Cortes and N. Lawrence and D. Lee and M. Sugiyama and R. Garnett},
 pages = {},
 publisher = {Curran Associates, Inc.},
 title = {Galileo: Perceiving Physical Object Properties by Integrating a Physics Engine with Deep Learning},
 volume = {28},
 year = {2015}
}

@article{degrave2019differentiable,
  title={A differentiable physics engine for deep learning in robotics},
  author={Degrave, Jonas and Hermans, Michiel and Dambre, Joni and Wyffels, Francis},
  journal={Frontiers in neurorobotics},
  volume={13},
  pages={6},
  year={2019},
  publisher={Frontiers Media SA}
}

@InProceedings{Yang_2025_CVPR,
    author    = {Yang, Kun and Liu, Yuxiang and Cui, Zeyu and Liu, Yu and Zhang, Maojun and Yan, Shen and Wang, Qing},
    title     = {NTR-Gaussian: Nighttime Dynamic Thermal Reconstruction with 4D Gaussian Splatting Based on Thermodynamics},
    booktitle = {Proceedings of the Computer Vision and Pattern Recognition Conference (CVPR)},
    month     = {June},
    year      = {2025},
    pages     = {691-700}
}

@inproceedings{wangetgs,
  title={ETGS: Explicit Thermodynamics Gaussian Splatting for Dynamic Thermal Reconstruction},
  author={Wang, Zhongwen and Ling, Han and Zhang, Weihao and Sun, Yinghui and Sun, Quansen},
  booktitle={The Fourteenth International Conference on Learning Representations}
}

@INPROCEEDINGS{7299096,
  author={Saponaro, Philip and Sorensen, Scott and Kolagunda, Abhishek and Kambhamettu, Chandra},
  booktitle={2015 IEEE Conference on Computer Vision and Pattern Recognition (CVPR)}, 
  title={Material classification with thermal imagery}, 
  year={2015},
  volume={},
  number={},
  pages={4649-4656},
  doi={10.1109/CVPR.2015.7299096}}

@article{wang2021neus,
  title={NeuS: Learning Neural Implicit Surfaces by Volume Rendering for Multi-view Reconstruction},
  author={Wang, Peng and Liu, Lingjie and Liu, Yuan and Theobalt, Christian and Komura, Taku and Wang, Wenping},
  journal={arXiv preprint arXiv:2106.10689},
  year={2021}
}

@incollection{icml2020_2086,
 author = {Gropp, Amos and Yariv, Lior and Haim, Niv and Atzmon, Matan and Lipman, Yaron},
 booktitle = {Proceedings of Machine Learning and Systems 2020},
 pages = {3569--3579},
 title = {Implicit Geometric Regularization for Learning Shapes},
 year = {2020}
}

@inproceedings{marchingcubes,
author = {Lorensen, William E. and Cline, Harvey E.},
title = {Marching cubes: A high resolution 3D surface construction algorithm},
year = {1987},
isbn = {0897912276},
publisher = {Association for Computing Machinery},
address = {New York, NY, USA},
url = {https://doi.org/10.1145/37401.37422},
doi = {10.1145/37401.37422},
booktitle = {Proceedings of the 14th Annual Conference on Computer Graphics and Interactive Techniques},
pages = {163–169},
numpages = {7},
series = {SIGGRAPH '87}
}

@misc{waseem2025physicsinformedneuralnetworksthermophysical,
      title={Physics-Informed Neural Networks for Thermophysical Property Retrieval}, 
      author={Ali Waseem and Malcolm Mielle},
      year={2025},
      eprint={2511.23449},
      archivePrefix={arXiv},
      primaryClass={cs.LG},
      url={https://arxiv.org/abs/2511.23449}, 
}

@manual{ansys,
  title        = {Ansys Mechanical User's Guide},
  author       = {{ANSYS, Inc.}},
  organization = {ANSYS, Inc.},
  year         = {2024},
  note         = {Ansys Mechanical, Release 2024 R2, Help System},
  url          = {https://ansyshelp.ansys.com/public/Views/Secured/corp/v242/en/wb_sim/ds_Home.html}
}

\begin{appendices}
\newpage
\section{Dataset Description}
\label{app:data}
The general experimental settings are summarized as follows:
\begin{itemize}
    \item The initial object temperature is set to $22\,^{\circ}\mathrm{C}$ ($295.15\,\mathrm{K}$).
    \item Material properties, including density, specific heat capacity, thermal conductivity, convective heat-transfer coefficient, and surface emissivity, are assigned according to the object-specific values listed in Table~\ref{tab:materials}.
    \item Cooling, heating, warming, and held-out testing conditions are defined on a per-scene basis, as summarized in Table~\ref{tab:scenes}.
    \item For cooling and warming scenes, no external heating is applied ($\dot{Q}=0$), whereas the applied heating power in active heating scenes varies by object.
    \item The ambient temperature is not fixed across all experiments, but is specified separately for each scene, ranging from $-15\,^{\circ}\mathrm{C}$ to $100\,^{\circ}\mathrm{C}$ in the evaluated configurations.
\end{itemize}

\begin{table}[htbp]
\centering
\caption{Material, surface, and boundary properties of the evaluated objects.
$\rho$ denotes density, $c_p$ specific heat capacity, $k$ thermal conductivity, $h$ convective heat-transfer coefficient, and $\epsilon_s$ surface emissivity.}
\label{tab:materials}
\small
\setlength{\tabcolsep}{4pt}
\begin{tabular}{lllllll}
\hline
Object & Material & $\rho$ & $c_p$ & $k$ & $h$ & $\epsilon_s$ \\
& & kg/m$^3$ & J/(kg$\cdot$K) & W/(m$\cdot$K) & W/(m$^2\cdot$K) & -- \\
\hline
Cube & Oak & 935.7 & 1685 & 0.4528 & 3.00 & 0.90 \\
Cylinder & Stainless Steel & 7750 & 480.0 & 15.10 & 5.00 & 0.30 \\
Sphere & Copper & 8960 & 385.0 & 398.0 & 5.00 & 0.30 \\
Bunny & ABS & 1030 & 1440 & 0.1997 & 3.00 & 0.92 \\
Bear & Glass & 2465 & 898.6 & 1.003 & 3.00 & 0.85 \\
Car & Aluminum Alloy & 2770 & 875.0 & 148.6 & 5.00 & 0.20 \\
\hline
\end{tabular}

\vspace{2pt}
\raggedright
\footnotesize{ABS, acrylonitrile butadiene styrene.}
\end{table}

\begin{table}[t]
\centering
\caption{Scene configurations for training and testing. Each object has two training scenes and two held-out testing scenes. $T_a$ denotes the ambient temperature in $^\circ$C, and $\dot{Q}$ denotes the applied heating power in W.}
\label{tab:scenes}
\scriptsize
\setlength{\tabcolsep}{4pt}
\begin{tabular}{l*{8}{c}}
\toprule
\multirow{3}{*}{Object}
& \multicolumn{2}{c}{Train - Cooling}
& \multicolumn{2}{c}{Train - Heating}
& \multicolumn{2}{c}{Test - Warming}
& \multicolumn{2}{c}{Test - Heating} \\
\cmidrule(lr){2-3} \cmidrule(lr){4-5} \cmidrule(lr){6-7} \cmidrule(lr){8-9}
& $T_a$ & $\dot{Q}$
& $T_a$ & $\dot{Q}$
& $T_a$ & $\dot{Q}$
& $T_a$ & $\dot{Q}$ \\
\midrule
Cube & 0.0 & 0.0 & 22.0 & 1.0 & 50.0 & 0.0 & 22.0 & 1.0 \\
Cylinder & -15.0 & 0.0 & 22.0 & 20.0 & 100.0 & 0.0 & 22.0 & 10.0 \\ 
Sphere & 0.0 & 0.0 & 22.0 & 10.0 & 50.0 & 0.0 & 22.0 & 5.0 \\
Bear & -15.0 & 0.0 & 22.0 & 10.0 & 100.0 & 0.0 & 22.0 & 10.0\\
Car & -15.0 & 0.0 & 22.0 & 50.0 & 100.0 & 0.0 & 22.0 & 150.0 \\
\bottomrule
\end{tabular}
\end{table}

\subsection{Data Processing Pipeline}
We first load the common (reconstructed) surface meshes and convert into a solid body. Afterwards, the triangular surface representations are extended to quaratic tetrahedral elements each with 4 vertices and 6 mid-edge nodes. Due to the post-processing of the reconstructed mesh, the geometric points and observataiont inherently own offsets, thereby requiring a projection from observed temperatures to geometric mesh.

\subsection{Thermal Infrared Video Rendering}
In this section, we describe the procedure for rendering multi-view thermal infrared video sequences from a given STL mesh and corresponding pseudo ground-truth temperature evolutions, used for both training and evaluation. The STL mesh serves as the geometric input to ANSYS transient heat simulations, from which per-vertex temperature profiles are obtained. These temperature evolutions are treated as ground truth, capturing dynamic thermal behaviors under warming, cooling, and heating scenarios.

Let the sparse thermal point cloud be defined as
\begin{equation}
\mathcal{P}=\{(\mathbf{x}_i, \mathbf{T}_i)\}_{i=1}^{N},
\qquad
\mathbf{x}_i \in \mathbb{R}^3,\; \mathbf{T}_i = [T^0, \ldots, T^    t]^T \in \mathbb{R}^{t},
\end{equation}
where \(\mathbf{x}_i\) denotes the 3D position of the \(i\)-th sample and \(\mathbf{T}_i\) its associated temperature time series over $t$ steps.

We describe the rendering process for a single frame at time step \(t\), which generalizes to all frames in the sequence. For a given camera view, each thermal point is transformed from world to camera coordinates as
\begin{equation}
\mathbf{x}_i^c = \mathbf{H}_{cw}\mathbf{x}_i,
\end{equation}
where \(\mathbf{H}_{cw}\) denotes the world-to-camera transformation. Under the pinhole projection model, the intrinsic matrix is defined as
\begin{equation}
\mathbf{K} =
\begin{bmatrix}
f_x & 0 & c_x \\
0 & f_y & c_y \\
0 & 0 & 1
\end{bmatrix},
\end{equation}
where \(f_x\) and \(f_y\) denote the focal lengths, and \((c_x, c_y)\) is the principal point. The projected pixel coordinates are then given by
\begin{equation}
u_i = f_x \frac{x_i^c}{z_i^c} + c_x,
\qquad
v_i = f_y \frac{y_i^c}{z_i^c} + c_y.
\end{equation}

\begin{figure}[t]
    \centering
    \includegraphics[width=\linewidth]{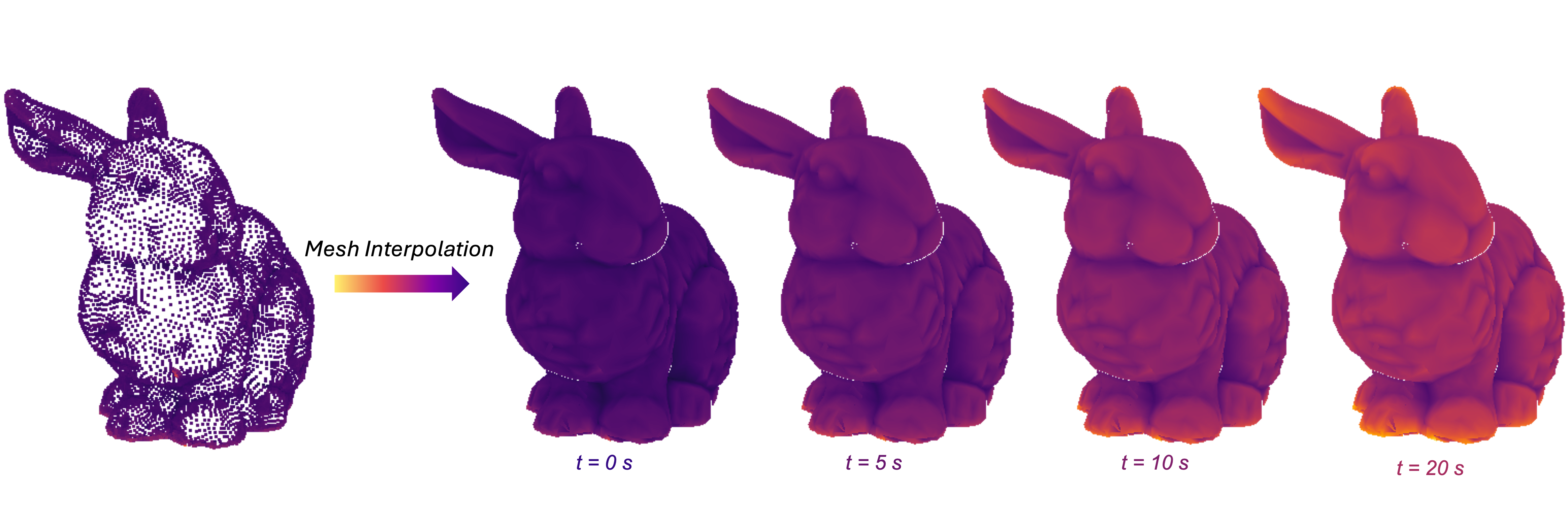}
    \caption{Thermal rendering pipeline. From left to right: sparse thermal samples are first projected onto the mesh surface, followed by vertex-based interpolation to produce dense thermal renderings over all time steps.}
    \label{fig:thermal_rendering_pipeline}
\end{figure}

To ensure geometric consistency, only visible thermal samples are retained. Let \(D(u,v)\) denote the rendered depth map of the scene. A projected thermal point is considered visible if
\begin{equation}
\left| z_i^c - D(u_i,v_i) \right| < \varepsilon,
\end{equation}
where \(\varepsilon\) is a depth tolerance parameter. This step produces \textbf{a sparse set of temperature observations in the image} while suppressing occluded samples.

Since the projected thermal samples are typically sparse, a dense thermal field is obtained by transferring temperature values onto the mesh surface. The STL surface is represented as a triangular mesh
\begin{equation}
\mathcal{M}=(\mathcal{V},\mathcal{F}),
\end{equation}
where \(\mathcal{V}=\{\mathbf{v}_m\}_{m=1}^{M}\) is the set of mesh vertices and \(\mathcal{F}\) is the set of triangular faces. For each mesh vertex \(\mathbf{v}_m\), the temperature is interpolated from its \(k\) nearest thermal samples using inverse-distance weighting,
\begin{equation}
\hat{T}_m = \sum_{j \in \mathcal{N}_k(\mathbf{v}_m)} w_{mj} T_j,
\end{equation}
where \(\mathcal{N}_k(\mathbf{v}_m)\) denotes the set of \(k\) nearest thermal points to \(\mathbf{v}_m\). The interpolation weights are defined as
\begin{equation}
w_{mj} =
\frac{d_{mj}^{-1}}
{\sum_{\ell \in \mathcal{N}_k(\mathbf{v}_m)} d_{m\ell}^{-1}},
\qquad
d_{mj} = \|\mathbf{v}_m - \mathbf{x}_j\|_2.
\end{equation}

Once vertex temperatures are assigned, the temperature at a pixel \(\mathbf{p}\) inside a visible triangle \((\mathbf{v}_{m_1}, \mathbf{v}_{m_2}, \mathbf{v}_{m_3})\) is computed by barycentric interpolation,
\begin{equation}
\hat{T}(\mathbf{p}) =
\lambda_1(\mathbf{p}) \hat{T}_{m_1}
+ \lambda_2(\mathbf{p}) \hat{T}_{m_2}
+ \lambda_3(\mathbf{p}) \hat{T}_{m_3},
\end{equation}
subject to
\begin{equation}
\lambda_1(\mathbf{p}) + \lambda_2(\mathbf{p}) + \lambda_3(\mathbf{p}) = 1,
\qquad
\lambda_r(\mathbf{p}) \geq 0,\; r \in \{1,2,3\}.
\end{equation}

Finally, the interpolated thermal field is normalized for visualization,
\begin{equation}
\bar{T}(\mathbf{p}) =
\mathrm{clip}
\left(
\frac{\hat{T}(\mathbf{p}) - T_{\min}}
{T_{\max} - T_{\min}},
0, 1
\right),
\end{equation}
where \(T_{\min}\) and \(T_{\max}\) define the temperature range. The normalized scalar field is then mapped to a color space to produce the final thermal image. This formulation combines sparse thermal observations with geometry-aware surface interpolation, generating dense thermal renderings that remain consistent with both scene geometry and visibility constraints.

\section{Supplementary Methods}
\label{app:method}

\subsection{SDF-based geometry reconstruction}
\label{app:sdf}

To recover object geometry from multi-view RGB observations, we adopt the signed-distance-function (SDF) representation of NeuS~\cite{wang2021neus}. The scene is modeled by two functions: a signed distance function $f(\mathbf{x})$, which maps a spatial position $\mathbf{x}\in\mathbb{R}^{3}$ to its signed distance from the object surface, and a color function $c(\mathbf{x},\mathbf{v})$, which predicts the RGB appearance of a point $\mathbf{x}$ viewed from direction $\mathbf{v}$. The reconstructed surface is defined as the zero level set of the SDF,
\begin{equation}
\mathcal{S}=\{\mathbf{x}\in\mathbb{R}^{3}\mid f(\mathbf{x})=0\}.
\end{equation}

The geometry and appearance networks are optimized through differentiable volume rendering. For each camera ray $\mathbf{r}$ emitted from camera origin $\mathbf{o}$ along viewing direction $\mathbf{v}$, sampled points are generated as
\begin{equation}
\mathbf{x}_i=\mathbf{o}+t_i\mathbf{v},
\qquad i=1,\dots,n,
\end{equation}
with $t_{i+1}>t_i>0$. The rendered color is computed by accumulating the contribution of all sampled points,
\begin{equation}
\hat{C}(\mathbf{r})=\sum_{i=1}^{n} w_i \hat{c}_i,
\qquad
w_i=T_i\alpha_i,
\end{equation}
where $\hat{c}_i$ is the predicted color at the $i$-th sample point, $T_i=\prod_{j=1}^{i-1}(1-\alpha_j)$ is the accumulated transmittance, and $\alpha_i$ is the opacity induced by the SDF-based rendering formulation of NeuS~\cite{wang2021neus}.

The reconstruction objective combines a color reconstruction term and an eikonal regularization term,
\begin{equation}
\mathcal{L}_{\mathrm{surface}}
=
\mathcal{L}_{\mathrm{color}}
+
\lambda_{\mathrm{eik}}\mathcal{L}_{\mathrm{eikonal}},
\end{equation}
where
\begin{equation}
\mathcal{L}_{\mathrm{color}}
=
\sum_{\mathbf{r}\in\mathcal{R}}
\left\|
\hat{C}(\mathbf{r})-C(\mathbf{r})
\right\|_1
\end{equation}
measures the discrepancy between rendered and observed RGB values, and
\begin{equation}
\mathcal{L}_{\mathrm{eikonal}}
=
\sum_{\mathbf{x}\in\mathcal{X}}
\left(
\|\nabla f(\mathbf{x})\|_2-1
\right)^2
\end{equation}
encourages the network to behave as a valid signed distance function~\cite{icml2020_2086}.

After optimization, the object surface is extracted from the learned SDF using marching cubes~\cite{marchingcubes}. The framework is not restricted to NeuS specifically; alternative high-quality surface reconstruction methods can be substituted as long as they produce a metrically scaled surface suitable for tetrahedralization.




\subsection{Physical quantities}
\label{app:units}

Table~\ref{tab:units_heat_app} summarizes the physical quantities appearing in the transient heat-transport model.

\begin{table}[h]
\centering
\caption{Physical quantities in the transient heat-transport model.}
\label{tab:units_heat_app}
\begin{tabular}{lll}
\toprule
Property & Description & Unit \\
\midrule
$T$ & Temperature & $\mathrm{K}$ \\
$\rho$ & Density & $\mathrm{kg/m^3}$ \\
$c_p$ & Specific heat capacity & $\mathrm{J/(kg\cdot K)}$ \\
$k$ & Thermal conductivity & $\mathrm{W/(m\cdot K)}$ \\
$\dot{q}$ & Volumetric heat generation rate & $\mathrm{W/m^3}$ \\
$h$ & Convection coefficient & $\mathrm{W/(m^2\cdot K)}$ \\
$\varepsilon$ & Surface emissivity & -- \\
$\sigma$ & Stefan--Boltzmann constant & $\mathrm{W/(m^2\cdot K^4)}$ \\
$\alpha$ & Thermal diffusivity, $k/(\rho c_p)$ & $\mathrm{m^2/s}$ \\
$\beta$ & Scaled convection coefficient, $h/(\rho c_p)$ & $\mathrm{m/s}$ \\
$\gamma$ & Scaled emissivity term, $\varepsilon/(\rho c_p)$ & $\mathrm{s\,K/J}$ \\
\bottomrule
\end{tabular}
\end{table}

\subsection{Differentiable finite-element heat-transfer solver}
\label{app:jaxfem}

The transient heat-transfer problem is solved on the reconstructed tetrahedral mesh using JAX-FEM, a differentiable finite-element framework implemented in JAX. Let $\Omega$ denote the reconstructed object volume and $\partial\Omega$ its boundary. The governing equation is
\begin{equation}
\rho c_p \frac{\partial T}{\partial t}
=
\nabla\cdot(k\nabla T)+\dot{q}
\qquad \text{in } \Omega,
\end{equation}
with Robin-type boundary conditions on $\partial\Omega$,
\begin{equation}
-k\nabla T\cdot\mathbf{n}
=
h(T_s-T_{\infty})
+
\varepsilon \sigma
\left(
T_s^4-T_{\mathrm{env}}^4
\right)
-
q_b.
\end{equation}

In the implementation, the volumetric conduction term is discretized over tetrahedral elements, and the surface exchange term is evaluated over boundary quadrature points. Time integration is performed implicitly, so that the temperature field at the current time step is obtained from the previous step by solving a nonlinear residual equation.

For numerical stability, the radiative term is evaluated in linearized form around the previous time step. Specifically, the effective radiative coefficient is computed using the previous surface temperature, yielding a Robin-type update that remains stable under implicit time integration while preserving differentiability of the overall simulation pipeline.

External heating is introduced as a boundary heat flux applied to selected boundary quadrature points. When heating is active only over a prescribed time interval, the corresponding boundary flux is enabled only for time steps satisfying $t<t_{\mathrm{heat}}$.

A key advantage of this formulation is that the forward solution operator is differentiable with respect to the unknown thermophysical fields. In practice, the element residual assembly, boundary-kernel evaluation, implicit time stepping, and solution operator are all embedded in the JAX autodifferentiation pipeline. Gradients of the observation loss can therefore be backpropagated through the entire transient solve.

\subsection{Linear and quadratic finite-element discretizations}
\label{app:linear_quadratic}

ThermoField supports both linear and quadratic tetrahedral discretizations. In the linear setting, temperature is represented on first-order tetrahedral elements (TET4). In the quadratic setting, the mesh is promoted to second-order tetrahedral elements (TET10), which introduce mid-edge nodes and provide more accurate interpolation of the temperature field.

In practice, inverse optimization is performed in two stages. The first stage uses the linear discretization to obtain a stable coarse solution at lower computational cost. The second stage initializes from the linear-stage parameters and refines the solution on the quadratic discretization. This staged strategy provides a favorable trade-off between optimization stability and final simulation accuracy. The linear stage reduces the tendency of the inverse problem to fit fine-scale numerical artifacts prematurely, whereas the quadratic stage better resolves the spatial structure of the temperature field and heat fluxes on complex geometry.

\subsection{Spatial field parameterization and interpolation}
\label{app:field_param}

Thermophysical quantities are represented as neural fields on sparse surface control points. Let $\{p_m\}_{m=1}^{M}$ denote sampled control points on the reconstructed surface. Their coordinates are first normalized to the bounding box of the reconstructed geometry, and then enriched by positional encoding,
\begin{equation}
\psi(p_m)=
\left[
\tilde{p}_m,\,
\|\tilde{p}_m\|_2,\,
\sin(2^0\pi\tilde{p}_m),\,
\cos(2^0\pi\tilde{p}_m),\,
\dots,\,
\sin(2^{L-1}\pi\tilde{p}_m),\,
\cos(2^{L-1}\pi\tilde{p}_m)
\right],
\end{equation}
where $\tilde{p}_m$ is the normalized coordinate and $L$ is the number of positional-encoding frequencies.

For each trainable field $\phi$, a multilayer perceptron maps $\psi(p_m)$ to a scalar output. The output is passed through a sigmoid function and rescaled to a prescribed physical interval $[\phi_{\min},\phi_{\max}]$,
\begin{equation}
\phi(p_m)
=
\phi_{\min}
+
(\phi_{\max}-\phi_{\min})
\,\sigma(g_{\phi}(\psi(p_m))),
\end{equation}
where $g_{\phi}$ is the neural field and $\sigma(\cdot)$ denotes the logistic sigmoid.

The predicted control-point values are interpolated to boundary vertices using precomputed control-point neighborhoods and interpolation weights. They are then extended to tetrahedral cells through face-based interpolation on the reconstructed boundary surface. This produces consistent control-point, boundary-vertex, and cell-wise field values for each thermophysical quantity.

When only a subset of the surface is thermally active, control-point sampling can be biased toward high-activity surface regions, and field values outside the active region can optionally be fixed to their initialization values. These mechanisms improve data efficiency and reduce unnecessary field flexibility in weakly observed regions.

\subsection{Observation-space supervision}
\label{app:obs_interp}

Thermal observations are provided as temperature samples on the reconstructed surface over time. During preprocessing, each observed surface sample is assigned to its nearest boundary triangle on the reconstructed mesh, and barycentric weights are computed relative to that triangle. During optimization, simulated nodal temperatures on the boundary mesh are interpolated back to the observation points using the corresponding triangle shape functions. The observation loss is therefore evaluated directly at the measured surface samples rather than on an auxiliary regular grid.

\subsection{Inverse objective and regularization}
\label{app:loss}

Let $T_s^{\mathrm{obs}}(\mathbf{x}_i,t)$ denote the observed surface temperatures and $\hat{T}_s(\mathbf{x}_i,t;\Theta)$ the simulated temperatures generated from field parameters $\Theta$. The data term is
\begin{equation}
\mathcal{L}_{\mathrm{data}}
=
\frac{1}{N}
\sum_{i,t}
\left\|
\hat{T}_s(\mathbf{x}_i,t;\Theta)-T_s^{\mathrm{obs}}(\mathbf{x}_i,t)
\right\|_2^2.
\end{equation}

To discourage non-physical local oscillations in the recovered fields, we impose a smoothness penalty over neighboring surface control points,
\begin{equation}
\mathcal{L}_{\mathrm{smooth}}
=
\sum_{\phi\in\mathcal{F}}
\frac{1}{|\mathcal{N}|}
\sum_{(p_i,p_j)\in\mathcal{N}}
\left(
\phi(p_i)-\phi(p_j)
\right)^2,
\end{equation}
where $\mathcal{F}$ denotes the set of trainable thermophysical fields and $\mathcal{N}$ denotes neighboring control-point pairs. This term promotes spatial coherence of the recovered fields and reduces the tendency of the inverse problem to explain observations through fragmented local excursions.

We further apply standard parameter regularization to the neural-field parameters,
\begin{equation}
\mathcal{L}_{\mathrm{reg}}
=
\|\Theta\|_2^2.
\end{equation}
The final objective is
\begin{equation}
\mathcal{L}
=
\mathcal{L}_{\mathrm{data}}
+
\lambda_{\mathrm{smooth}}\mathcal{L}_{\mathrm{smooth}}
+
\lambda_{\mathrm{reg}}\mathcal{L}_{\mathrm{reg}}.
\end{equation}

Empirically, the smoothness term is important for controlling the width of the recovered field distribution. Weakening or removing this term typically leads to substantially broader 5--95\% field widths, indicating that the optimizer increasingly explains the observed temperatures through local compensation rather than through a physically coherent thermophysical field.

\subsection{Optimization details}
\label{app:opt}

The neural thermophysical fields are optimized with Adam. Each trainable field may use its own learning-rate scaling factor, while the full objective is evaluated jointly over one or more thermal scenarios for the same object. When multiple thermal sequences are available, their losses are averaged during each optimization epoch so that complementary thermal processes constrain the same underlying thermophysical field. Model checkpoints, field snapshots, and training histories are recorded throughout optimization for later evaluation and analysis.





\end{appendices}



\end{document}